# Machine Learning and Visualization in Clinical Decision Support: Current State and Future Directions

Gal Levy-Fix[a,1], Gilad J. Kuperman[b], Noémie Elhadad[a]

*a Biomedical Informatics; Columbia University; New York, NY, USA*

*b Memorial Sloan Kettering Cancer Center; New York, NY, USA*

**Abstract**

Deep learning, an area of machine learning, is set to revolutionize patient care. But it is not yet part of standard of care, especially when it comes to individual patient care. In fact, it is unclear to what extent data-driven techniques are being used to support clinical decision making (CDS). Heretofore, there has not been a review of ways in which research in machine learning and other types of data-driven techniques can contribute effectively to clinical care and the types of support they can bring to clinicians. In this paper, we consider ways in which two data driven domains—machine learning and data visualizations—can contribute to the next generation of clinical decision support systems. We review the literature regarding the ways heuristic knowledge, machine learning, and visualization are – and can be – applied to three types of CDS. There has been substantial research into the use of predictive modeling for alerts, however current CDS systems are not utilizing these methods. Approaches that leverage interactive visualizations and machine-learning inferences to organize and review patient data are gaining popularity but are still at the prototype stage and are not yet in use. CDS systems that could benefit from prescriptive machine learning (e.g., treatment recommendations for specific patients) have not yet been developed. We discuss potential reasons for the lack of deployment of data-driven methods in CDS and directions for future research.

[1] Corresponding author
 *Email address:* gf2308@cumc.columbia.edu (Gal Levy-Fix)

## 1. Introduction

Learning health systems hold the promise for providing more personalized, higher quality, safer, and efficient care.[1] The learning health systems pipeline involves systematically gathering clinical data, learning from that data and generating evidence, and feeding it back to clinicians in real-time to help with decision making. This process highly depends on the robust development, evaluation, and adoption of clinical decision support (CDS) in clinical care to deliver knowledge to the point of care. However, for CDS to help realize the goals of a learning health system, numerous challenges have to be addressed. Challenges to the effective use of CDS include not being sufficiently patient-specific, utilizing simplistic CDS logic, lacking generalizability, and failing to address human factor issues.[2] Leveraging the recent developments in machine learning and data visualization, especially in combination, could help overcome some of these challenges. Machine learning (ML) methods have the potential to enhance CDS tools by generating new knowledge from gathered data, providing better patient specificity, supporting the identification of complex patterns, and improving generalizability to different patients and conditions. Data and information visualization (dataVis) techniques, from static to interactive visualizations to more complex visual analytics, have the potential to assist with feeding back information to clinicians and improve the interpretability and transparency of CDS systems. Machine learning and data visualization provide complimentary benefits to CDS and may be synergistic in combination (**Figure 1**). Thus, there is a strong case for greater focus on leveraging machine learning and data visualization in combination to help the realization of a learning health system.

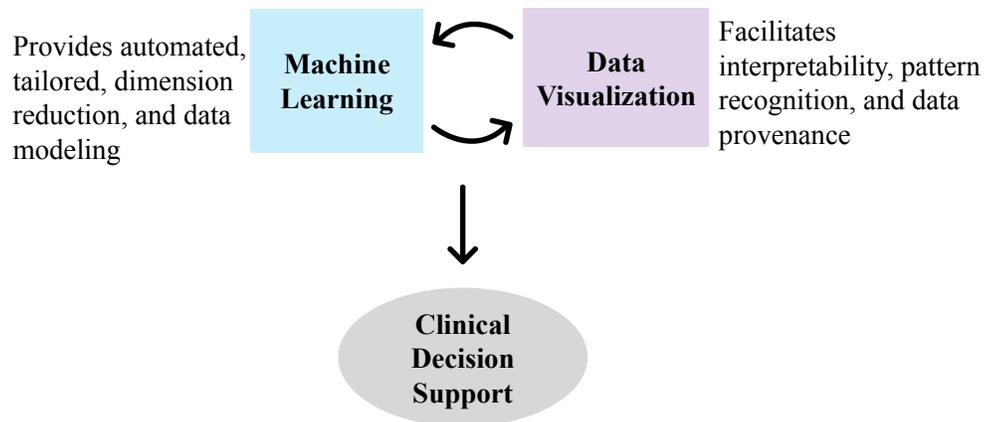

**Figure 1.** Synergy of data visualization, machine learning, and clinical decision support. This review is dedicated to describing and synthesizing the current state of the literature on machine learning and data visualization methods used for clinical decision support. The review also identifies lessons learned and points out opportunities to use machine learning and data visualization to improve CDS and to further the goals of a learning health system. The target audience for this review include informaticians, designers, machine learning experts, and practitioners interested in reading how synergies between machine learning and visualization could address the challenges of clinical decision systems.

## 2. Materials and Methods

This review is based on a survey of the CDS literature and literature describing methodology developed for CDS applications. Pool of papers was compiled through a search on PubMed for the terms "clinical decision support", "machine learning", or "visualization." In addition, papers were compiled from machine learning for health conference proceedings (Machine Learning for Healthcare, NeurIPS Machine Learning 4 Health Workshops) and IEEE Visualization conference proceedings. We focused our search to publications from 2010 to 2019, but also included earlier seminal works dating back to 1959. Additional papers were identified through "pearl growing," until we reached thematic saturation. We restricted our focus to papers describing clinician-facing clinical decision support, whether patient-specific or cohort-level, and utilizing EHR data collected through clinical documentation. In final count of papers included in this review is 244. Papers were classified into three general types of CDS, although lessons learned from analysis of current work in these CDS types should also generalize to other CDS types. The three CDS types, as identified by Musen and colleagues,[3] are referred to in this review as *Infobutton*, *Content Aggregation and Organization*, and *Alert*.

- **Infobuttons** are a type of a CDS developed to help clinicians retrieve external resources relevant to the care of their patients such as scientific publications and guidelines. As medical evidence is constantly generated and updated and as clinicians have less time at the point of care, Infobuttons make it easier to stay up to date and well informed.

- **Content Aggregation and Organization (CAO)** CDS is used to re-organize and present patient-level or cohort-level information to clinicians in a way that facilitates understanding, pattern recognition, and decision making. Current Electronic Health Record (EHR) systems contain large amounts of data even for single patients, making the tasks of information gathering and synthesis cognitively difficult and time consuming. CAO CDS aims to help centralize and crystalize patient data available for better and easier decision making.

- **Alert** CDS provides alerts, reminders, and recommendations in the context of patient data, clinician actions (such as medication orders), and clinical knowledge. Due to the high volume of data available in the EHR, limited clinician time, and evolving medical guidelines, clinicians may miss important information regarding a given patient that could lead to better and safer care. Alert CDS produces a single output such as a prediction, an alert, or a set of recommendations to clinicians in order to help direct action and prevent medical errors.

The following section of the review synthesizes previous work on each CDS type (Infobuttons, CAO CDS, and alert CDS) and the machine learning and data visualization methods utilized. Literature on each CDS type is grouped and described by the type of methods they utilize (**Figure 2**). We review how each CDS type has applied: **(1)** heuristics-based knowledge development methods (heuristics) defined to be expert curated rules or knowledge-based sources such as ontologies; **(2)** machine learning (ML) defined to be data-driven and learning-based methods for knowledge development; and **(3)** data visualization (dataVis) defined to be the advanced visual representation of data and information using static or interactive graphs, diagrams, or pictures to convey information; and **(4)** any combination of these three methods.

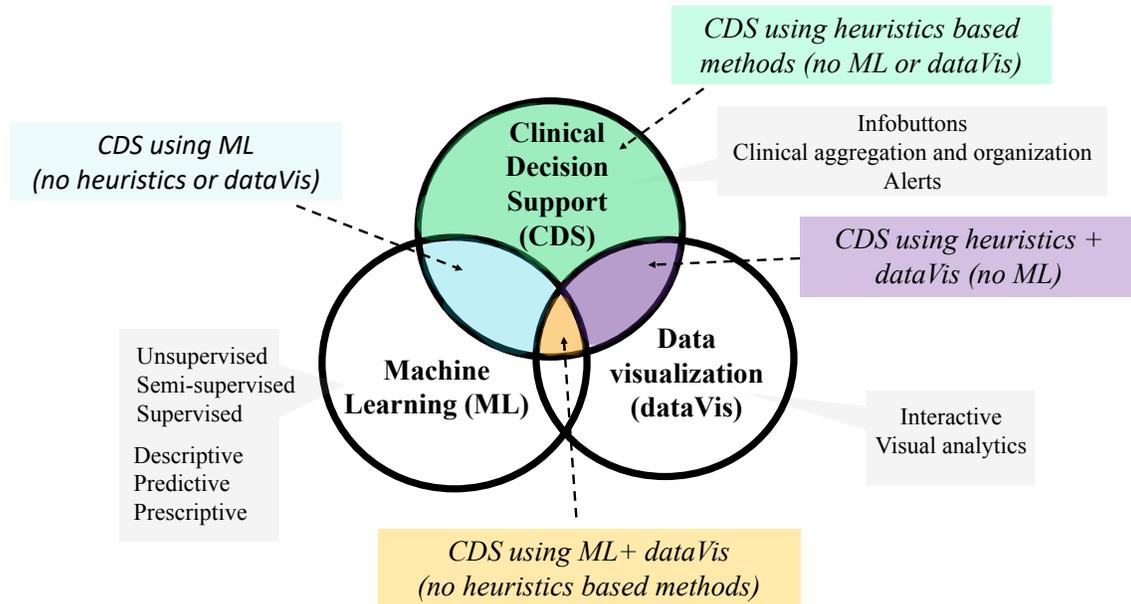

**Figure 2.** Venn diagram showcasing the intersections between clinical decision support, machine learning, and data visualization. We refer to heuristics-based methods as rules that are expert-curated or that rely on knowledge sources such as ontologies. Machine-learning methods include clinical data-driven methods. Visualization methods include static, interactive, as well as advanced visual analytics from clinical data.

## 3. Results

The following section synthesizes the reviewed literature, identifying methods utilized in current CDS work in machine learning and data visualization. Sections 3.1, 3.2, and 3.3 are dedicated to each CDS type in turn, each including a table with methodological examples utilized in the CDS type . A discussion of research gaps and opportunities in how machine learning and visualization can support that CDS type is also present.

### 3.1 Infobutton Clinical Decision Support

Infobuttons are systems developed to help clinicians retrieve external resources such as scientific publications and guidelines that are relevant to their patients and informational needs. As medical evidence is constantly being generated and updated and with clinicians having less available time, Infobuttons make the task of accessing up to date medical evidence relevant to their clinical cases easier.

### 3.1.1 Types of Methods for Infobutton CDS

Of the three CDS types, Infobuttons represent the most common CDS implemented in the EHR.[3] Research on heuristics-based Infobuttons, with most work taking place in the 1990s, leverages a combination of ontological knowledge and rules to identify clinical concepts in patient records and construct relevant search queries to look for relevant resources in scientific article databases or online.[4–10]

Infobuttons leveraging machine learning largely focus on the personalization and summarization of the outside resources retrieved by the system and returned to clinicians. These systems largely represent experimental stand-alone systems that have not been integrated in EHR systems, as have traditional Infobuttons. General approaches of these works include context aware scientific article summarization, recommendation of outside resources based on patient data, learning to rank models of articles based on clinician search queries, and question-answering related to a patient's clinical case.[11–16]

Very limited works have leveraged data visualization for Infobutton CDS in isolation or in combination with machine learning. The limited work in this area has looked at an interactive citation screening system for improved clinical question answering.[17]

**Table 1.** Examples of Infobutton CDS by method type

| Papers | Type of method utilized | Description |
| --- | --- | --- |
| Powsner et al. 1989 [5] | Heuristics | Utilize rule-based Medline searched by clinical topic |
| Cimino et al. 1997 [9] | Heuristics | Use terminology knowledge (Medical Entities Dictionary) to select queries and resources |
| Elhadad et al. 2005 [11] | ML | Use Natural language processing to tailor summaries of scientific articles based on the clinical context of the patient. Evaluated vis user study of simulated clinical task compared effectiveness of tailored summary, to non-tailored summary and general article search |
| Monteiro et al. 2015 [12] | ML | Recommender system of reports and studies based on patient information and clinical context |
| Donoso-Guzmán& Parra 2018 [17] | ML+dataVis | Compare two relevance feedback algorithms, Rocchio and BM25, in an interactive visualization for citation screening. Evaluated efficiency and effectiveness of tool in citation screening in user group |
| More papers by method category: | Heuristics | [4,6,10,18,19] |
| | ML | [14,15] |

### 3.1.2 Gaps and Opportunities for ML and dataVis in Infobutton CDS

Results of heuristics-based Infobuttons may still return large amounts of content for clinicians to review in order to find relevant information for their patients. For instance, scientific literature about a specific clinical concept might return hundreds of highly relevant publications. Furthermore, most Infobuttons search resources for one piece of information in the patient record and does not consider combination of clinical concepts, potentially reducing the relevance and usefulness of the retrieved sources. One way to reduce the complexity and size of results is to curate content based on clinical expertise, but this might limit the scope of Infobuttons as well as their sustainability as new evidence emerges.

Data-driven methods can be used to organize further the results of Infobuttons, whether to condense and synthesize the evidence or to personalize and tailor the evidence to the clinician's information needs and clinical context, thus making the information search quicker and more efficient for the clinician.[20] An additional promise for data-driven Infobuttons, rather than rule-driven ones is increased generalizability to different types of searches and concepts with less reliance on manual curation of content. Supervised solutions, however still require annotated datasets which in the clinical context can be time-consuming and expensive to obtain. It is also important to note that data-driven systems have so-far been mostly evaluated for accuracy and effectiveness in a laboratory setting outside of a deployed, real-world setting. More research is required to evaluate their utility and performance in the clinical setting.

The lack of integration of visualization in this line of research is also a missed opportunity. Work outside of the health domain has shown that data visualization can help users identify relevant information in information retrieval tasks and facilitate thematic analysis of large sets of documents.[21–24]

### 3.2 Clinical aggregation and organization (CAO) Clinical Decision Support

CAO clinical decision support systems either re-organize or summarize patient information. Dashboards, that select specific data points and presents them in a centralized way, or summarizers which synthesize entire records belong to these types of CDS. These systems aim to help users with information that is difficult to digest in its original form in the EHR due to its volume, complexity, or it scattered nature in the EHR.

### 3.2.1 Type of methods for CAO decision support systems

Literature on CAO decision support largely leverage heuristics-based methods such as expert curated variable selection and knowledge-based sources to organize [25,26] or summarize the information.[27–35] These tools have mainly focused on extractive summaries [36] which extract selected information from the patient record into condensed tables,[27–33] with fewer works providing abstractive summaries[37] which reformulates the patient's data, largely to automatically infer patient problem lists using structured data.[35,37,38]

Several data visualization technique have also been proposed in combination with heuristics-based CAO systems. Popular approaches for visual extractive summaries have been small visuals and patient data temporal views.[39–52]A few examples also exist of visualizations of abstractive summaries or reorganization of selected patient data.[53–63] Machine learning used for CAO systems have largely focused on generating abstractive summaries of the patient rather than extractive summaries. That is, reducing patient data dimensionality and complexity into more salient, condensed, and digestible form. These approaches have included automatically generating short narrative description of patients' data and generating the patient's problem list using natural language processing methods (NLP) and supervised machine learning methods.[64–71] Another machine learning approach that has been used to reducing patients' data dimensionality is data-driven phenotyping. Although most commonly proposed for features engineering in predictive tasks, interpretable abstraction of patient's clinical data from data driven phenotyping could also be used for patient summarization in clinical decision support.[72] Computational approaches that propose data-driven phenotyping include probabilistic models,[72–77] deep learning,[78–83] clustering,[84] and decision trees.[85]

Few examples in the literature have used a combination of machine learning and visualization methods for CAO systems. One group of works focus on leveraging machine learning and interactive visualization to showcase cohort visualizations aimed to assist clinicians with patient-level decision making. These works mostly divide into performing two tasks: 1) computation of patient sequence similarity using different clustering methods [86–88]; and 2) frequent patterns identification using advance association rules and latent model methods.[89–91] Another group of work leverage the machine learning and visualization for patient-level visualization. These works largely focus on generating abstractive summaries of patients' data using semi-supervised and unsupervised methods and visualizing those abstractions.[92,93] Evaluation methods utilized for systems leveraging machine learning and visualization include usability studies by clinical experts,[86,88,89,91] interactivity performance,[86] and prediction performance using patient-level abstractions. [92,93]

**Table 2.** Examples of CAO Clinical decision support by method type.

| Papers | Type of method utilized | Description |
|---|---|---|
| Alkesic et al. 2017 [26] | Heuristics | Organize clinical content using manual tagging of EHR content for chronic disease tracking |
| Meystre & Haug 2006 [34] | Heuristics | Infer patient problems using knowledge-based sources |
| Powsner & Tufte 1994 [39] | dataVis | Patient record summary using small graphs showing laboratory results, medications, vitals, and imaging. |
| Bui et al. 2007 [53] | dataVis | Problem centric patient record temporal abstractive summary using knowledge-based source |
| Van Vleck & Elhadad 2010 [69] | ML | Natural language processing and classification to predict problem relevance for clinical summarization. Automated patient problem summaries compared to expert generated gold standard |
| Joshi et al. 2016 [94] | ML | Learning identifiable patient phenotypes using non-negative matrix factorization. Qualitative evaluation of clinical expert of learned phenotypes and performance in mortality prediction |
| Guo et al. 2018 [91] | ML+dataVis | Use tensor decomposition to identify latent evolutions of care sequences. Present threads of latent sequences in treatment sequences |
| Joshi et al. 2012 [93] | ML+dataVis | Utilize novel clustering algorithm to generate layered-grouping of patient states. Real time visual of patient severity by organ system during ICU stay |
| More papers by method category: | Heuristics | [25,27,28,30–35] |
| | ML | [66,67,69–76,84,85,95–98] |
| | dataVis | [39–50,52–58,60–63,99–101] |
| | ML+dataVis | [86–88,90,92] |

*3.2.2 Gaps and Opportunities for ML and dataVis in CAO Clinical Decision Support*

Heuristics-based CAO systems have been shown to improve physicians' information retrieval capabilities, reduce information overload, improve patient outcomes, and guideline compliance.[25,27,29,30,102] However, these systems mostly focus on extractive summaries which may still contain overwhelming amount of information.[103] Furthermore, a lack visualization use can limit in the effectiveness of the proposed summaries.[104]

Introduction of machine learning methods, especially those that are unsupervised and high-throughput,[83,105–107] automate dimensionality reduction of complex patient data into abstractive summaries that utilize more information from the patient record relative to the extractive summaries with little or no human input. However, few works in this area have been investigated specifically for CAO systems and often do not provide output that is geared for use by clinicians.

The use of data visualization have been shown to support pattern identification across patient parameters and time.[108] While visual summaries of patients' raw data preserves data provenance which can strengthen trust in the visuals,[109] they are limited in how many dimensions they can show [110] and may still lead to information overload.[103] Furthermore, previously proposed systems in this category have mostly been non-interactive which limit the capacity of the user to conduct exploratory analysis.[111] These systems fall short according to the Visual Information Seeking Mantra: Overview first, Zoom and Filter then Details-on-Demand.[112]

Works that combine both machine learning and data visualization methods are able to bypass some of the limitations seen in systems that only leverage one such methodology. However, most works leveraging both machine learning and data visualization methods have focused on cohort-level visualizations rather than patient-level visualizations.[86,88–91,113] Furthermore, like for data-driven Infobuttons, few of these systems have been evaluated for clinical utility. Methods outside of the health domain that can inform future research include automatic visual summaries of temporal new stories and topic modeling.[114,115]

### 3.3 Alert Clinical Decision Support

Alert clinical decision support produces a focused output such as a prediction about a specific outcome, an alert, or a set of recommendations to clinicians in order to help direct action and prevent medical errors in the context of patient data.

### 3.3.1 Type of methods for Alert CDS

Of the three clinical decision support types, alert CDS has the most sustained interest in the literature. Early work on these systems date back to the late 1950's and continued with a recent surge. Similar to the Infobutton systems, most mature systems implemented and used by clinicians today leverage knowledge sources and expert curated rules.[5,116–120] Heuristics-based CDS have largely underutilized visualization techniques. Existing examples of the use of visualization showcase patient data alongside the alert or recommendation.[121,122] Other work have proposed the use of visualization for knowledge base maintenance at the backend of alert systems but not for the use of clinicians.[123]

By contrast, the bulk of recent published work has focused on developing machine-learning methods that have the potential to assist in future alert CDS. Proposed machine learning methods have tackled a wide range of CDS applications and have leveraged a diverse set of approaches (**Figure 3**). Applications of machine learning methods developed for use in future alert CDS systems comprise disease and disease-stage prediction, optimal treatment prediction, and readmission and mortality prediction. The most popular machine learning approaches explored in recent years include deep learning and probabilistic methods.

Only a few systems leverage both machine learning and visualization. Systems that do utilize both methods motivate the use of visualization for added interpretability, model transparency, data provenance, and usability.[91,95,124–126]

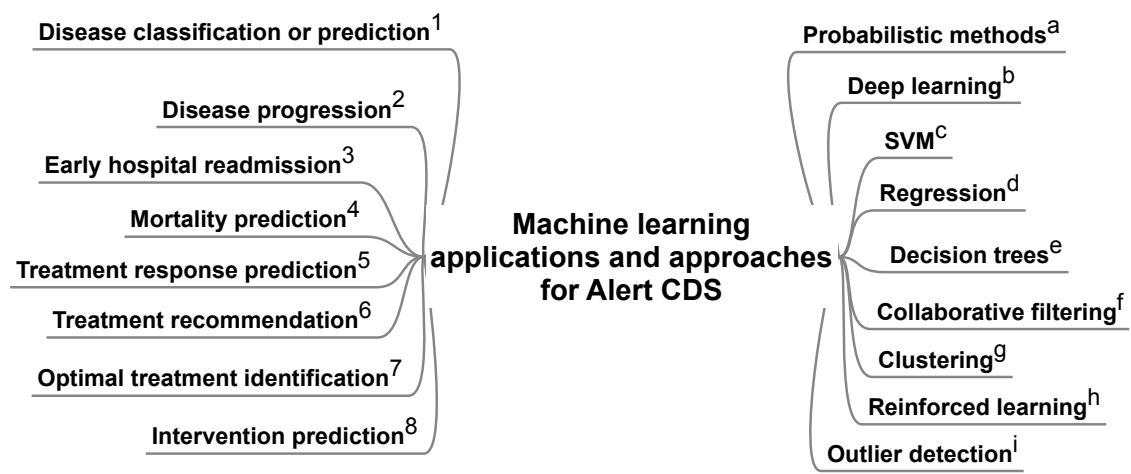

**Figure 3.** Machine learning applications and approaches for alert CDS. Applications include disease classification or prediction,[83,98,106,123,127–172] disease progression, [84,137,143,154,166,172–185] hospital readmission,[186–188] mortality prediction,[94,188–191] treatment-response prediction,[80,90,107,192–197] treatment recommendation,[196–202] treatment identification,[177,205–209] and intervention prediction,[196, 203, 208–211]. Approaches include probabilistic methods,[128,134,141,143,152,164,178,184,185,205,210,214–219] deep learning, [80,130,134,153,155,157,160,161,166–168,171,173,175,180,183,204,207,208,220] support vectors,[134,142,150,164,170,173,175] regression,[129,136,150,156,159,179] decision trees,[147,150] collaborative filtering,[154] clustering,[193,221] reinforcement learning,[36,209,222] and outlier detection [223].

**Table 3.** Examples of alert CDS by method type

| Papers | Type of method utilized | Description |
|---|---|---|
| Warner et al. 1972, Warner 1979; Kuperman et al. 1991 [116,120,224] | Heuristics | Rule-based logical operators to assist with diagnosis |
| Miller et al. 1982; 1989[119,225] | Heuristics | Knowledge-based system that can construct and resolve differential diagnoses. Evaluated for accuracy compared to human experts. Evaluated for clinical utility |
| Goldstein et al. 2000; Gennari et al. 2003 [121,122] | Heuristics+ dataVis | Guidelines and ontology-based treatment recommendation system for chronic disease. presents the patients raw data related to the chronic problem such as the patient's blood pressure readings over time |

| Warner et al. 1964 [128] | ML | Use Bayes' Theorem to the diagnosis of congenital heart disease. Compared accuracy of system to that of clinical experts |
|---|---|---|
| Wang et al. 2014 [174] | ML | Use unsupervised probabilistic model to model disease progression. |
| Tsoukalas et al. 2015 [95] | ML+dataVis | Partially observable markov decision process model. interactive graphical interface for optimal treatment for Sepsis. Includes visual of patient vital history over time, state transition probabilities, patient state history, and optimal action. Evaluate generalized error of approach and in external tasks of mortality prediction and length of stay prediction |
| Jeffery et al. 2017 [124] | ML+dataVis | Mobile app to showcase the predicted probability of cardiac arrest overtime, including forecasted risk for the next 24 hours. Evaluate tool for usability in a lab setting with target audience |
| More papers by method category: | Heuristics: | [117,118] |
| | ML | [127,129–133,136–140,106,141,206,174,142–152,173,175–177,192,193,197,199,218,154–160,163,178,154,179,180,219,80,195,196,198,210,211,207,161,220,164–172,214,181–185,226,194,107,208,209,205,212,213,201–204,83,98] |
| | ML+dataVis | [125,126,227] |

### 3.3.2 Gaps and Opportunities for ML and dataVis in Alert CDS

Heuristics-based alert CDS have been found to improve healthcare processes, but that there is still little robust evidence of leading to improvements in clinical outcomes, costs, workload and efficiencies.[228,229] Commonly cited limitations of heuristics-based alert systems pertain to their narrow clinical focus, most likely due to the need for manual curation of clinical expertise in the systems. Few systems are 'high-throughput' or able to assist on wide range of conditions and patient types. In practice, this can translate in multiple CDS systems, each relevant to a specific subset of patients, with a need to deploy and manage them each to support diverse types of patients and clinical contexts. This can lead to 'alert overload', with too many systems firing alerts to clinicians, each with little awareness of the others.

Adding data visualizations to heuristics-based alert CDS can help with interpretability and data provenance, leading to higher confidence in the system and usability. However very few works have explored this research space.

Introducing machine-learning techniques into alert CDS can help generate evidence directly from gathered clinical data, reducing the need for clinical knowledge to be codified manually by experts.[106,131,138] Moreover, machine learning methods can also handle many more predictors and complex relationships such as non-linearity, interactions, and temporality that would be hard to codify in knowledge-based systems.[138,176,230]

Machine learning methods can also handle data with missingness, sparsity, noise, and irregular sampling.[226,231,232] However, machine learning methods intended for CDS have often been criticized as uninterpretable, prone to data biases, and dependent on the data they are evaluated on.[145,233–235] This can make the comparison of models problematic when evaluated on different data and also be regarded as 'too risky' to incorporate into clinical decision making. Other significant limitations of data-driven alert CDS is their lack of alignment with clinical workflows, with few proposed methods evaluating clinical utility with clinically meaningful metrics, and they have not been deployed to clinical settings.[236,237] For instance, some approaches which ignore when data are generated in the clinical workflow, can lead to data leakage when predicting outcomes and would not be possible to implement.

While alert CDS that introduce data visualization for the end user are often more mindful of clinical workflow they attempt to support, they too have largely been evaluated on model accuracy, face validity of visualization, and interface usability in a laboratory setting.[124,126,177,227,238] The need for interpretable and transparent learning methods has also been recognized outside of the health domain. Several reports have cited the integration of data visualization for the interpretation and understanding of machine learning methods and their results as key.[239,240]

## 4. Discussion and directions for future work

Implemented CDS have largely utilized heuristic methods such as knowledge sources and expert-driven rules to deliver decision support to clinicians. Machine learning and visualization techniques have been leveraged to various degrees depending on the type of CDS. Some work in machine learning has been used for Infobuttons; CAO systems have integrated data visualization techniques for information presentation, and a large body of work exists on machine learning methods that can contribute to alert CDS.

Opportunities abound for the expanded use of machine learning and data visualization methods for clinical decision support. Reviewed work suggests data-driven approaches can be effectively leveraged for robust information abstraction whether for more tailored information retrieval, outcome prediction, or patient record summarization. Visualization has been shown to provide better representation and interpretability for users. Used in combination, machine learning and advanced data visualization could help unleash the full potential of next-generation clinical decision support. We identify the following key items that are critical to investigate to translate their success into practice.

- **Bringing advanced techniques into clinical practice.** Although a fair amount of research has been dedicated to innovative CDS tools and methods, bringing them into clinical practice is still an outstanding challenge. It is commonly reported that developed tools are mismatched with the actual clinical workflow they are meant to support.[236,241] This sentiment is echoed for many CDS types but mostly regarding analytic tools meant to assist with clinical decision making (i.e., statistical or machine-learning tools).[236] Close collaborations between researchers and clinical partners early on and not just at the evaluation stage may help remedy this disconnect. Motivating new innovative methods and systems with actual clinical needs can lead to higher adoption rates of new CDS in clinical practice.

  Another open challenge to bringing CDS research into practice is gaining practitioners' trust and fostering use of new visual analytics and ML-based CDS tools. More evidence regarding the effectiveness of such tools in clinical settings could sway sentiments in the right direction. Evaluation metrics for proposed CDS systems need to be ones that clinicians care about rather than benchmark metrics (such as high AUC scores) leveraged in unrealistic evaluation tasks and settings. Only after undertaking these steps can new CDS tools be adopted and potentially have positive impact on health of patients.

- **Aligning techniques towards an impact on care.** Machine learning research in the field has largely focused on predictive modeling. These models focus on predicting a single outcome given the data available, mostly at one point in time in a patient's health trajectory. Very few work investigate generating longer-term, more holistic trajectories of patients' potential states of health, with and without different interventions.[84,196,242,243] Future work should assure that clinical value will be garnered from such systems. As robustness and generalizability of data-driven techniques expand the realm of clinical decision support systems, further research should also examine whether other, not-yet explored clinical tasks can benefit from these techniques.

Moreover, beyond prediction tasks, characterization and descriptive tasks using learning techniques could propel CDS forward. The applications that describe and show what has happened to a patient in time can be helpful beyond CAO tools, and help reduce the complexity of EHR data, as a further way to help with interpretability and explainability of models. Finally, prescriptive modeling, while not yet investigated in the clinical domain, may prove critically useful at the point of care. Such tools could propose action recommendations and optimal treatment options. For these tools to be effective, they would require high levels of trust from clinicians. Trust could be fostered using data visualization methods for greater interpretability and model transparency as well as rigorous evaluation for clinical utility.

- **Moving from CDS to learning health systems.** Several works using machine learning and data visualization have been dedicated to large cohort analysis. While these can be useful when managing large panels of patients, most day-to-day work of clinicians pertains to caring for individual patients. There is a lot to gain if previously proposed cohort-level tools could be adapted to provide personalized insights for individual patients at the point of care.

  In the reverse direction, future work should examine if care for individuals can inform guidelines applied to population's health. Currently, medical knowledge is largely integrated into CDS through the manual codification of guidelines. Previous work has shown that these guidelines are not always adhered to, not always available, and may become out of date.[244] For a truly learning health system, collected data needs to be analyzed to generate insight and up-to-date knowledge that is then fed back into the health system through CDS for clinicians to use.

## 5. Conclusion

To work towards a learning health system, CDS systems need to play a major role. To do so they need to be able to learn from gathered data, assist in generating new insight, showcase results to practitioners, gain the trust of users, and be well adapted to clinical workflows. This review demonstrates that the complimentary nature of machine learning and visualization methods for CDS can further these goals. While machine learning and visualization have been integrated in various types of CDS their combination is still an open and promising research direction.

## 6. Acknowledgements

This work was supported by NLM T15 LM007079 (GL) and NSF award #1344668 (NE).

**Conflict of interest**

No conflict of interest was declared.